\documentclass{article}

    \usepackage[nonatbib, preprint]{neurips_2019}

\usepackage[utf8]{inputenc} % allow utf-8 input
\usepackage[T1]{fontenc}    % use 8-bit T1 fonts
\usepackage{hyperref}       % hyperlinks
\usepackage{url}            % simple URL typesetting
\usepackage{booktabs}       % professional-quality tables
\usepackage{amsfonts}       % blackboard math symbols
\usepackage{nicefrac}       % compact symbols for 1/2, etc.
\usepackage{microtype}      % microtypography

\usepackage{float}
\usepackage{graphicx}
\usepackage{subfig}

\usepackage{cite}

\title{Greedy Bandits with Sampled Context}

\author{
  \textbf{Dom Huh} \\ 
  $^{1}$Department of Electrical and Computer Engineering, George Mason University, Virginia, USA \\
  \texttt{dhuh4@gmu.edu}
}

\begin{document}

\maketitle

\begin{abstract}
Bayesian strategies for contextual bandits have proved promising in single-state reinforcement learning tasks by modeling uncertainty using context information from the environment. In this paper, we propose Greedy Bandits with Sampled Context (GB-SC), a method for contextual multi-armed bandits to develop the prior from the context information using Thompson Sampling, and arm selection using an $\epsilon$-greedy policy. The framework GB-SC allows for evaluation of context-reward dependency, as well as providing robustness for partially observable context vectors by leveraging the prior developed. Our experimental results show competitive performance on the Mushroom environment in terms of expected regret and expected cumulative regret, as well as insights on how each context subset affects decision-making.
\end{abstract}

\section{Introduction}

Many real-world tasks, often requiring the balance of exploration and exploitation trade-off \cite{Sutton} through experimentation, can be formulated into a multi-armed bandit (MAB) paradigm. The MAB paradigm, in short, describes a system of $n$ machines, commonly referred to as arms, each with an unknown, often Bernoulli, distribution of the reward. For example, for personalized web-content recommendation tasks, an arm would represent the web content, and playing on that arm would represent advertising that web content. Thus, a MAB algorithm attempts to maximize the expected reward to reach the optimal reward, which is received from playing the optimal arms, based on past experimentation of different arms. Regret is the metric used to measure the difference between the expected reward to the optimal reward. Under the scope of the MAB paradigm, there exist three classes of MAB algorithm $\epsilon$-greedy \cite{Watkins}, upper confidence bound maximization \cite{Lai, Agrawal, Katehakis, Auer}, and uncertainty analysis under the Bayesian framework \cite{Bradt, Gittens, Thompson}. In this paper, we will focus on the latter approach, specifically utilizing Thompson Sampling \cite{Thompson, Chapelle}, a Bayesian strategy to optimize the MAB paradigm by probability matching using a Beta distribution, to handle context processing and $\epsilon$-greedy, a policy that chooses under the probability of $\epsilon$ random actions, and under the probability of 1-$\epsilon$ greedily, for arm selection.

Contextual MAB problems \cite{Li}, in contrast to vanilla MAB, takes into account the context which is presented at each trial, and often utilize past MAB algorithms with appropriate modification \cite{Li, Chu, Bietti, Riquelme}, representing each arm with a set of context features. These context features are often processed using linear or nonlinear models, or often an neural network.

In this paper, we introduce Greedy Bandits with Sampled Context (GB-SC), a contextual MAB framework using Thompson Sampling. To process the features, GB-SC treats each unique value of each context feature as a distribution, which are modeled using Thompson Sampling over $n$ trials. To select which arm to play, GB-SC will use formulate a confidence value, calculating the conditioning on the $k$ highest confidence samples of the context feature distribution, on each action in the action space and use $\epsilon$-greedy policy whether to play or not. 

\section{Preliminaries} \label{prelim}
To test our model, we experimented on the Mushroom environment specified in \cite{Guez, Blundell, Riquelme} using the Mushroom dataset from UC Irvine Machine Learning Repository \cite{Dua}. The Mushroom dataset includes 8124 examples of 23 species of mushrooms, each with 22 features and labeled poisonous or safe. The rules of the Mushroom environment is as follows: eating a safe mushroom provides reward of +5, eating a poisonous mushroom provides reward of +5 with probability 1/2 and reward of -35 otherwise, and no eating provides reward of 0. The set of actions contains whether to eat or not eat the mushroom. We ran 1500 trials to initially tune our algorithm before any tests. 

We diverge from the previous oracle defined in \cite{Guez, Blundell, Riquelme} to receive a reward +5 for a safe mushroom, or receive a reward 0 for a poisonous mushroom if the intended reward was -35, otherwise the reward remains +5. Thus, our oracle used to computed the optimal expected regret and the optimal cumulative expected regret acts as a better optimal policy than the one described in \cite{Guez, Blundell, Riquelme} as it takes into account the positive upside of taking risks of playing on a poisonous mushroom. From 100 independent 50 arms trials, the expected cumulative reward of our oracle is 60.24 points higher than the oracle in past works with a standard deviation of 16.46 points.

\section{Greedy Bandits with Sampled Context (GB-SC)}
In this section, we will first describe the context processing procedure, then the arm selection policy, and lastly formalize the Greedy Bandits with Sampled Context (GB-SC) algorithm in context with the Mushroom environment discussed in Section \ref{prelim}.

\begin{figure}[H]
  \centering{\includegraphics[width=\textwidth]{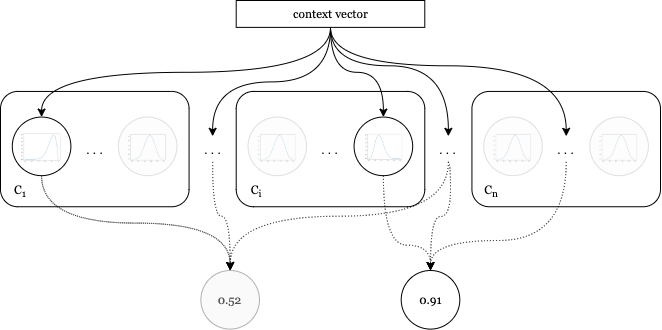}}
  \caption{Context Processing + Confidence: The context vector is split into $n$ context subsets, each activating only one node within its respective context subset. Each activated node is sampled, and $k$ samples with the highest confidence are used to compute the action confidence.}
  \label{context}
\end{figure}

Given the context from the environment, we can assume the context to be a set of discrete features, which we will refer to as context subsets. Within each context subset, we can model a random variable for each unique value. Thus, we have a set of random variables for each context subset. Assuming an action space consisting of 2 actions, we can utilize Thompson Sampling to model a Beta distribution for each random variable. In Fig. \ref{context}, the example illustrates context processing described here given a context vector with three explicitly shown context subsets. Further, the example shows activation of only a single node within each context subset.

With the activated nodes within the context subsets, we sample from their associated random variable to obtain a confidence, $C_i$. We determine confidence based on how close the sample, $S_i \sim Beta(\alpha_i,\beta_i)$, is from either zero or one. If the sample value is closer to 1, we assign the confidence to one action, and if the sample value is closer to 0, we assign the confidence to the other action.

\begin{equation}
    C_i = \left\{
        \begin{array}{ll}
            S_i & \quad S_i \geq 0.5 \\
            1 - S_i & \quad \textit{otherwise}
        \end{array}
    \right.
\label{confidence}
\end{equation}

Lastly, we average $k$ highest confidence for each action to obtain our action confidence, $C_{action}$. The choice of $k$ will be explored in Section \ref{results}.

\begin{equation}
    C_{action_i} = \frac{1}{k} \sum^{k}_{u}C_{i_u}
\label{action_confidence}
\end{equation}

For arm selection for $Arm_h$, we use the traditional framework of $\epsilon$-greedy policy, to determine whether to play $Arm_h$ or not. We set the exploration value, $\epsilon$, to the inverse of the current number of trials explored, $j$, to ensure a logarithmic regret, which was theoretically guaranteed in \cite{Auer}.

\begin{equation}
    Arm_h = \left\{
        \begin{array}{ll}
            \textit{argmax} C_{action} & \quad \epsilon \geq \frac{1}{j} \\
            X \sim \textit{Bern}(0.5) & \quad \textit{otherwise}
        \end{array}
    \right.
\label{arm}
\end{equation}

The GB-SC algorithm utilizes the context processing, confidence, and $\epsilon$-greedy policy to address the contextual MAB problem. With the Mushroom environment described in Section \ref{prelim}, GB-SC first creates 22 sets of random variables, each with distributions initialized to $Beta(\alpha,\beta)$, where $\alpha$ = 1 and $\beta$ = 1. GB-SC follows the update rule of Thompson Sampling specified in \cite{Thompson}, and additionally we scaled the update by the reward value. Thus, if the reward was positive, we updated the $\alpha$ parameter of that random variable by the reward value, and if the reward was negative, we updated the $\beta$ parameter of that random variable by the absolute reward value. The action confidence is computed by sampling the activated random variables and averaging the $k$ highest confidence for each action. The different values of $k$ was experimented in Section \ref{results}. If there isn't sufficient number of confidence for $k$ in an action, we simply use all confidence values. Then, GB-SC utilizes the $\epsilon$-greedy policy with with $\epsilon$ set to $\frac{1}{j}$ where $j$ is the current number of trials completed, to determine whether to play the arm or not. Playing an arm, in the context of the Mushroom environment, would mean whether to eat or not to eat a mushroom.

\section{Results} \label{results}
\begin{figure}[H]
  \centering{\includegraphics[width=\textwidth]{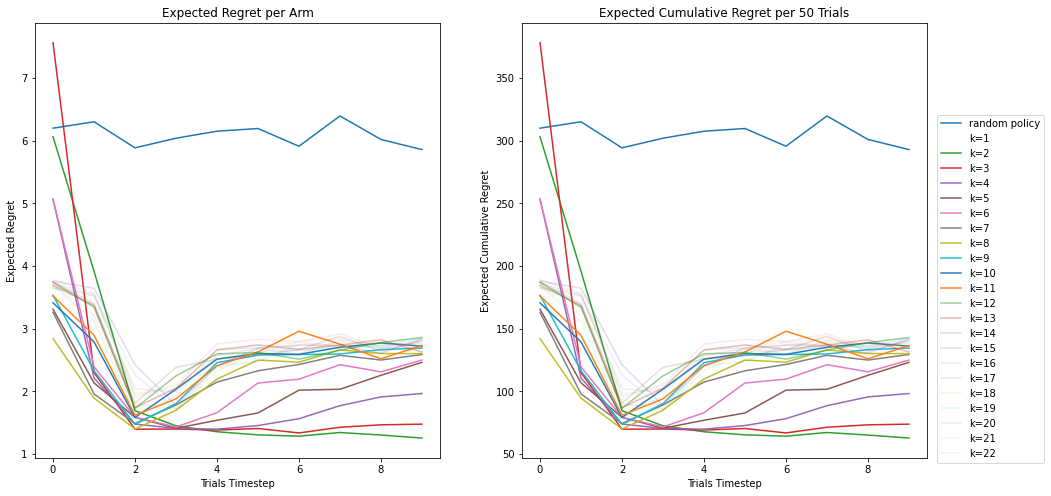}}
  \caption{Regret over Trials: The expected regret on each arm and the expected cumulative regret with 50 arm trials over 1500 trials at a time step of 150 trials for different $k$ values is shown as well as the regret using a random policy on this environment}
  \label{progress}
\end{figure}

As discussed in Section \ref{prelim}, we tested the GB-SC algorithm on the Mushroom environment, tracking the expected regret on each arm and the expected cumulative regret with 50 arm trials on the oracle we have defined. We used 1500 trials to obtain the final distribution for each context node, and progress can be seen in Fig. \ref{progress} for $\{k \| k \in N, 1 \leq k \leq22\}$, which represents the number of highest confidence values we take into account to calculate the action confidence. We updated the $\epsilon$ for the $\epsilon$-greedy policy every 150 trials of 50 arms, thus the time step of the Fig. \ref{progress} is at 150.

\begin{figure}[H]
  \centering{\includegraphics[width=\textwidth]{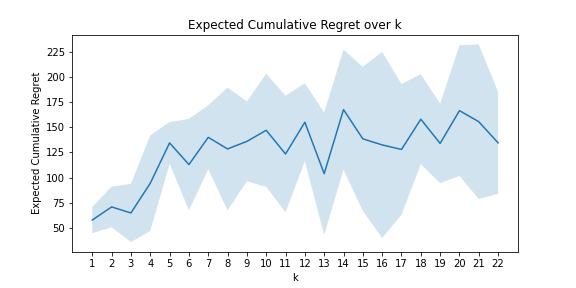}}
  \caption{Expected cumulative regret over $k$: The expected cumulative regret with 50 arm trials are measured after the 1500 trials. The measurements were taken with 10 independent trials, and the mean and standard deviation are shown.}
  \label{k_graph}
\end{figure}

We can see that there does exist a difference in how well model converges based on the $k$ value. So, we tested various values of $k$ and compared the performance in terms of expected cumulative regret on 50 arms with 10 trials in Fig. \ref{k_graph}. Based on Fig. \ref{k_graph}, we can conclude that the variance increases as $k$ increases, which could be that there are too many factors playing a role in the arm selection. Thus, the optimal values for $k$ seem to be between 1 and 3.

To gain insight on what features in the context vector are frequently used than the others, we can derive feature importance within the context vector on the Mushroom environment. To understand how to determine frequency, we explore two different approaches, and recommend other approaches to be experimented in further works. 

\begin{figure}[H]
  \centering{\includegraphics[width=0.9\textwidth]{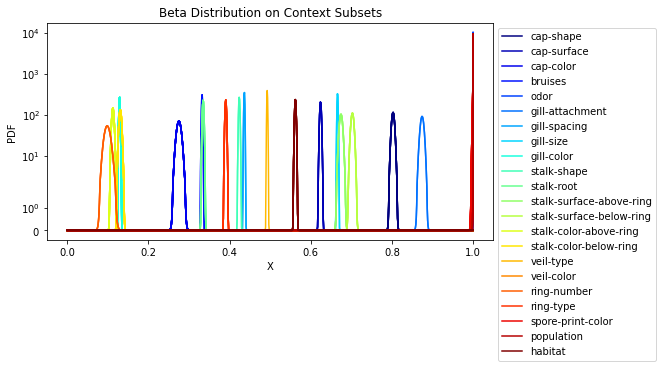}}
  \caption{Learned priors over Context Subsets: The probability density of the Beta distributions that represent the context subsets are shown on a symmetrical log scale.}
  \label{features}
\end{figure}

One option is to visualize all the distributions in each context subset and try to understand how each feature affects arm selection. This can be seen in Fig. \ref{features} using GB-SC with $k$=3. From Fig. \ref{features}, we can evaluate which context subset provide higher confidence values. To clarify, the closer the distribution is to one, the higher probability the context subset will be used for the action confidence for play, whereas the closer the distribution is to zero, the higher probability the context subset will be used for the action confidence for no play. However, this approach is not scalable for larger context vectors as the complexity of taking this approach and understanding how each context subset affects arm selection grows linearly with the number of context subsets and the number of unique values within each context subset. 

\begin{figure}[H]
  \centering{\includegraphics[width=0.9\textwidth]{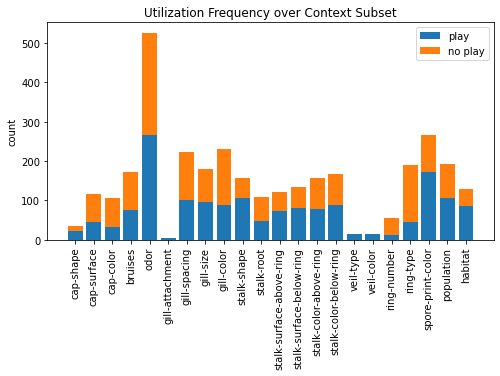}}
  \caption{Utilization Frequency over Context Subsets: The count frequency over each context subset is shown. The count is incremented by the number of arms used the context subset in 10 independent trials.}
  \label{utilization}
\end{figure}

Another option we explore is counting the number of times how frequently each context subset is selected for action confidence calculations for a given arbitrary $m$ number of trials. In Fig \ref{utilization}, we set $m$ = 10 using GB-SC with $k$=3, and we can see the most utilized context subsets to compute the action confidence, and also the least utilized context subsets as well. It is interesting to see that no context subset is never used, but some context subset were not used for no play.

\begin{figure}[H]
  \centering{\includegraphics[width=\textwidth]{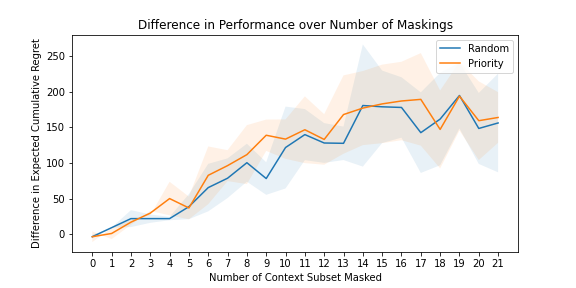}}
  \caption{Masking Context Subsets on Performance: The difference between expected cumulative regret over 10 independent trials between a masked context vector and non-masked context vector.}
  \label{nmasking}
\end{figure}

Additionally, we experimented with partially observable context vectors, where some context subsets would be missing. For this purpose, we should allocate a random variable in each context subset to handle nil values. For priority masking, we removed the more frequently utilized context subsets at a higher probability based on Fig \ref{utilization}, and for random masking, we set all the context subset to be masked all with equal probability. The results can be seen in Fig \ref{nmasking} using GB-SC with $k$=3.

\section{Conclusion}
In this paper, we have proposed the Greedy Bandit with Sampled Context(GB-SC) algorithm, a method of handling contextual MAB problem by using Thompson Sampling for context processing and $\epsilon$-greedy policy for arm selection. We demonstrate competitive performance compared to the baselines shown in \cite{Riquelme}, and we showed additional benefits such as insight into feature importance and being robust in arms with partially observable context vectors. In future works, we hope to expand the application with continuous context vectors, either through neural memory or some other method.

\bibliography{reference}{}

\end{document}